\documentclass[conference]{IEEEtran}
\usepackage{times}
\usepackage[numbers,sort&compress]{natbib}
\usepackage{multicol}
\usepackage{booktabs}
\usepackage{multirow}
\let\labelindent\relax
\usepackage{enumitem}
\usepackage{makecell}
\usepackage{graphicx}
\usepackage{cuted}
\usepackage{diagbox}
\usepackage{xurl}
\usepackage{tabularx}
\usepackage{colortbl}
\usepackage{xcolor}
\definecolor{hlgray}{RGB}{220, 220, 220}
\definecolor{hlblue}{RGB}{214, 235, 243}
\usepackage{amssymb}
\usepackage{amsmath}
\usepackage{subcaption}
\usepackage[bookmarks=true]{hyperref} 
\usepackage{titletoc}

\newcolumntype{Y}{>{\centering\arraybackslash}X}

\begin{document}

\title{Efficient and Reliable Teleoperation through Real-to-Sim-to-Real Shared Autonomy}

\author{
    \authorblockN{
    Shuo Sha$^{1}$,
    Yixuan Wang$^{1}$,
    Binghao Huang$^{1}$,
    Antonio Loquercio$^{2}$,
    Yunzhu Li$^{1}$
    }
    \authorblockA{
    $^{1}$Columbia University \quad $^{2}$University of Pennsylvania
    }
}

\maketitle

\begin{strip}
    \centering
    \vspace{-20pt}
    \includegraphics[width=\linewidth]{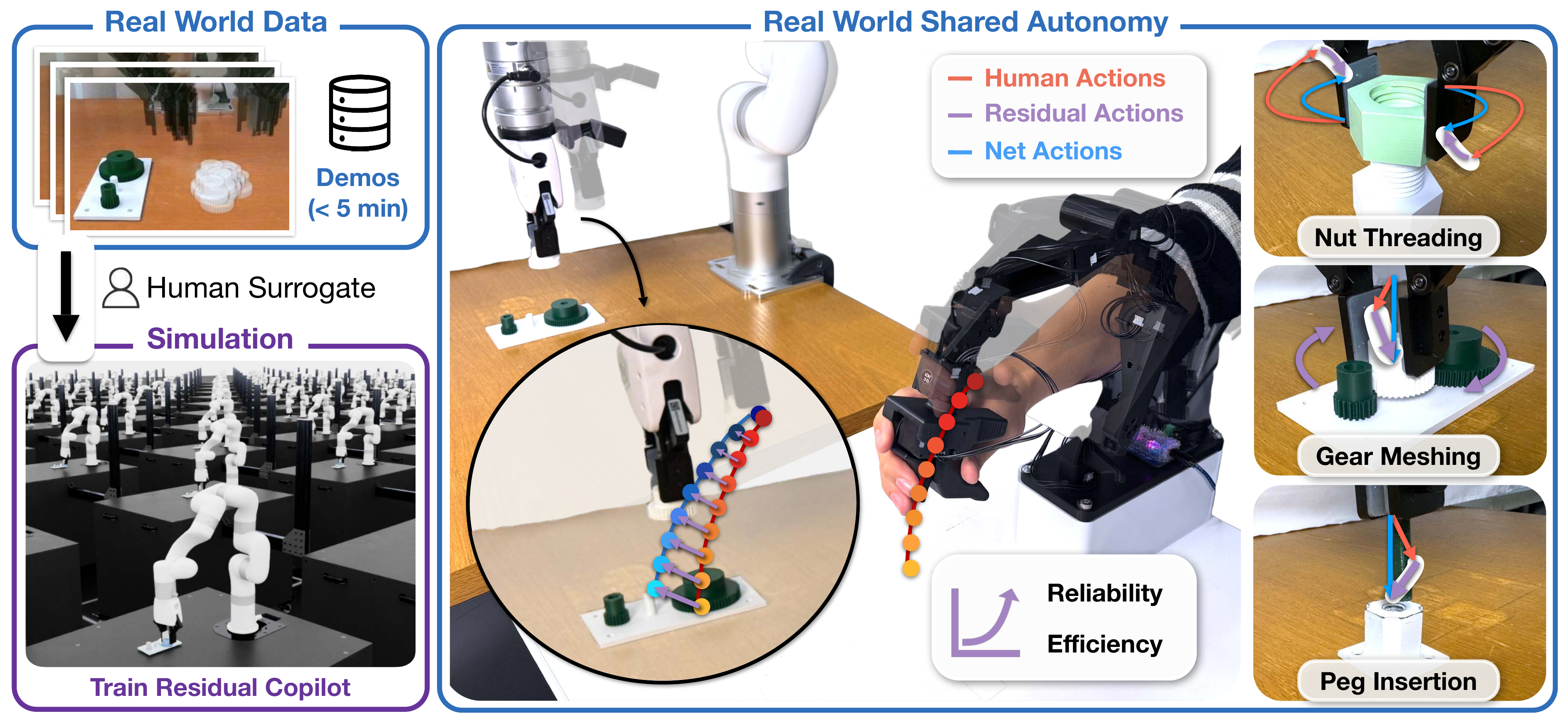}
    \vspace{-10pt}
    \captionof{figure}{\small
    \textbf{Overview of our real-to-sim-to-real shared autonomy framework.}
    A small amount of real teleoperation data ($<5$ minutes) is used to construct a lightweight human surrogate, which drives simulation-based training of a residual copilot policy. At deployment, the copilot provides low-level corrective actions that combine with human commands to produce reliable and efficient shared autonomy for fine-grained, contact-rich manipulation tasks, where direct teleoperation struggles with precise alignment, axis-constrained rotation, and contact regulation, including nut threading, gear meshing, and peg insertion.
    }
    \vspace{-10pt}
    \label{fig:teaser}
\end{strip}

\begin{abstract}
Fine-grained, contact-rich teleoperation remains slow, error-prone, and unreliable in real-world manipulation tasks, even for experienced operators. Shared autonomy offers a promising way to improve performance by combining human intent with automated assistance, but learning effective assistance in simulation requires a faithful model of human behavior, which is difficult to obtain in practice. We propose a real-to-sim-to-real shared autonomy framework that augments human teleoperation with learned corrective behaviors, using a simple yet effective k-nearest-neighbor (kNN) human surrogate to model operator actions in simulation. The surrogate is fit from less than five minutes of real-world teleoperation data and enables stable training of a residual copilot policy with model-free reinforcement learning. The resulting copilot is deployed to assist human operators in real-world fine-grained manipulation tasks. Through simulation experiments and a user study with sixteen participants on industry-relevant tasks, including nut threading, gear meshing, and peg insertion, we show that our system improves task success for novice operators and execution efficiency for experienced operators compared to direct teleoperation and shared-autonomy baselines that rely on expert priors or behavioral-cloning pilots. In addition, copilot-assisted teleoperation produces higher-quality demonstrations for downstream imitation learning. Website: \href{https://residual-copilot.github.io/}{\textcolor{magenta}{\texttt{https://residual-copilot.github.io/}}}
\end{abstract}

\section{Introduction}
Teleoperation is essential for executing complex robotic manipulation tasks and for collecting demonstrations to train visuomotor policies. However, for fine-grained, contact-rich, high-precision manipulation, teleoperation is often slow, error-prone, and unreliable. These limitations make teleoperation a practical bottleneck both for deploying robotic systems in the real world and for scaling high-quality demonstration data.

A key reason for this difficulty lies in a mismatch between human capabilities and the role assigned to them in conventional teleoperation. Humans excel at high-level intent specification and task reasoning, but struggle to provide continuous, low-level robustness through a teleoperation interface due to limited viewpoints, latency, and the embodiment gap. As a result, millimeter-scale alignment and contact regulation (crucial for precision manipulation) are poorly suited to direct human control and are better handled by an automated copilot.

Shared autonomy aims to address this mismatch by assisting human operators during task execution. Prior work~\cite{DexGen, DP_SA, MPC_SA, diffusion_copolicy, diffusion_safe, opt_4} has shown that shared autonomy can improve teleoperation by mapping human commands toward a stronger action prior, such as an expert policy, planner, or structured objective. A complementary line of work~\cite{DRL_SA, RL_SA_Drone, RPL_SA, RL_SA_budget} instead learns assistance directly as a copilot conditioned on the pilot command and system state, typically using model-free reinforcement learning or residual formulations that minimally adjust human input.

Despite their success, existing approaches face fundamental limitations. Methods that rely on expert priors shift the core challenge to obtaining the prior itself; once a competent autonomous policy exists, teleoperation is often no longer the limiting factor. Existing copilot-learning approaches avoid this assumption but either depend on extensive human-in-the-loop training, require large-scale data to learn an accurate human surrogate, or remain challenging to deploy in complex real-world manipulation settings.

In this work, we focus on copilot learning without assuming an expert prior. Rather than addressing teleoperation by first solving full autonomy, we treat autonomy as the harder problem and instead learn low-level corrective behaviors that preserve human intent. We propose a real-to-sim-to-real shared autonomy pipeline that trains a residual copilot policy using model-free reinforcement learning in simulation, driven by a lightweight k-nearest-neighbor (kNN) human surrogate fit from a small amount of real teleoperation data. By remaining within empirical data support, the kNN surrogate enables stable copilot training in simulation and effective transfer to the real world.
We evaluate our approach on a suite of fine-grained, contact-rich manipulation tasks, including nut threading, gear meshing, and peg insertion. Through simulation experiments and a user study with both novice and experienced operators, we show that our system improves task success for novice operators and execution efficiency for experienced operators compared to direct teleoperation and shared-autonomy baselines that rely on expert priors or behavioral-cloning pilots. In addition, our approach produces higher-quality demonstrations for downstream imitation learning.

In summary, our contributions are threefold: (1)~we propose a real-to-sim-to-real shared autonomy pipeline that assists teleoperation for a diverse suite of high-precision, contact-rich manipulation tasks; (2)~we show that a lightweight kNN model, constructed from less than five minutes of human demonstrations, can serve as an effective surrogate for a human during model-free policy training in simulation; and (3)~we demonstrate that our system increases the \emph{effectiveness} and \emph{efficiency} for both novice and experienced operators, enabling higher-quality demonstrations for imitation learning as well as practical assistive and remote teleoperation.

\section{Related Works}
\subsection{Shared Autonomy}
Shared autonomy combines human input with robot assistance during task execution and has been widely studied in teleoperation, surgical robotics, and assistive manipulation. To simplify terminology, we refer to the human user as the \textit{pilot} and the assistive agent as the \textit{copilot}.

Early work primarily focused on improving \emph{efficiency}, such as reducing time-on-task~\citep{min_time_1, min_time_2, casper, min_time_4, min_time_5, min_time_6, min_time_7, min_time_8, min_time_9, min_time_10, min_time_11, LILAC}, lowering input dimensionality~\citep{LA, min_input_2, min_input_3, min_input_4, min_input_5, min_input_6, min_input_7, min_input_8, min_input_9}, or enabling a single pilot to supervise multiple robots~\citep{max_output_1, max_output_2, max_output_3, max_output_4}. We focus instead on shared autonomy for high-precision, contact-rich manipulation, where improving the \emph{quality} of pilot input is essential.

\textbf{Expert-prior shared autonomy.}
One class of approaches assists teleoperation by blending pilot commands with an autonomous prior, such as a policy, planner, or action distribution. Assistance is often goal-conditioned, with pilot intent inferred explicitly or implicitly~\citep{goal_infer_1, goal_infer_2, goal_infer_3, goal_infer_4, goal_infer_5, goal_infer_6, goal_infer_7}. Pilot inputs are guided toward high-value or high-likelihood actions using Bayesian inference~\citep{bayes_infer_1, bayes_infer_3, bayes_infer_4}, inverse reinforcement learning~\citep{IRL, IRL_1, IRL_2, IRL_3, IRL_4, IRL_5}, or optimization-based formulations~\citep{opt_1, opt_2, opt_3, opt_4, MPC_SA}. Recent diffusion-based methods guide denoising with similarity gradients between pilot commands and policy outputs~\citep{DP_SA, DexGen, ITPS, diffusion_copolicy, diffusion_safe, consistent_diffusion_SA, interventional_diffusion}.

A key limitation of this class is its reliance on a competent autonomous prior. Acquiring expert data, training strong policies, or specifying accurate dynamics and objectives is often the dominant challenge. Moreover, prior work shows that blending pilot input with an autonomous policy does not guarantee the resulting behavior remains on the original motion manifold~\citep{ITPS, blending_critique, consistent_diffusion_SA}. These issues limit applicability where expert priors are unavailable or insufficiently robust.

\textbf{Copilot learning.}
A second class learns assistance directly as a copilot conditioned on the pilot command and system state. Representative approaches use reinforcement learning or POMDP formulations to trade off task return with deviation from pilot input~\citep{DRL_SA, RL_SA_budget, RL_SA_Drone, RL_SA_Driving, POMDP_Paper, RL_SA_Constraints}, while residual formulations improve sample efficiency by learning corrective actions on top of a nominal pilot policy~\citep{RPL_SA}. These methods avoid assuming an expert autonomous prior and are well suited to preserving human intent.

A central design challenge in copilot learning is the choice of pilot model during training. Prior work studies abstract pilots, such as noisy or laggy experts~\citep{DRL_SA, POMDP_Paper, RL_SA_budget, RL_SA_Constraints, RL_SA_Drone}, and demonstrates that pilot choice strongly affects generalization~\citep{DRL_SA, RPL_SA, RL_SA_budget}. More realistic pilots can be obtained via behavioral cloning from interaction data~\citep{RPL_SA, human_agent_study}, but parametric BC pilots are data-hungry and brittle under copilot-induced distribution shift in low-data regimes.

Our work builds on residual copilot learning and addresses this pilot-model bottleneck. Instead of fitting a parametric BC pilot, we use a lightweight k-nearest-neighbor (kNN) surrogate constructed from limited teleoperation data. By remaining within empirical data support, the kNN pilot enables stable copilot training in simulation and effective transfer to the real world. Unlike prior residual shared autonomy work~\citep{RPL_SA}, which primarily evaluates in simulation, we demonstrate real-world performance on fine-grained manipulation tasks.

\begin{figure*}[t]
    \centering

    \includegraphics[width=\linewidth]{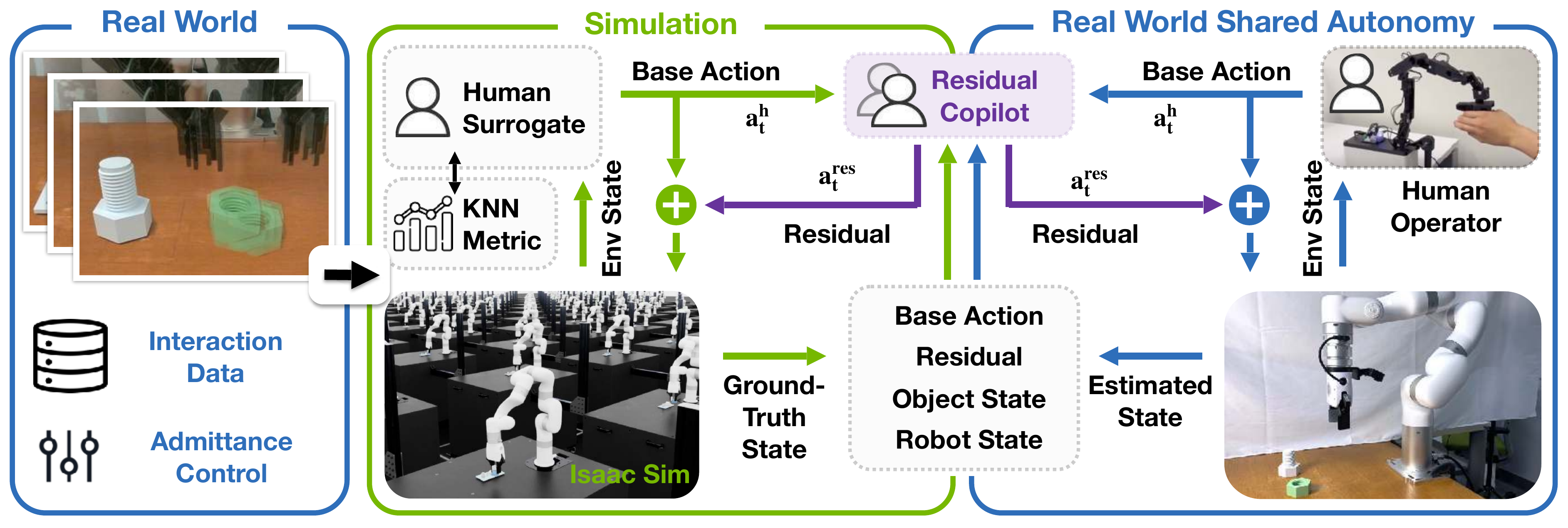}
    \vspace{-8pt}
    \captionof{figure}{\small
    \textbf{Method overview of residual copilot learning with a real-to-sim-to-real pipeline.}
    \textbf{Left:} less than five minutes of real-world teleoperation interaction data is collected under admittance control.
    \textbf{Center:} in simulation, a lightweight kNN-based human surrogate constructed from real-world data generates base actions conditioned on the environment state, while a residual copilot policy is trained with reinforcement learning to produce corrective residual actions.
    \textbf{Right:} at deployment, the residual copilot assists the human operator in zero-shot, conditioning on the estimated environment state and human base actions. The residual formulation preserves human intent while improving alignment, contact regulation, and execution robustness.
    }
    \label{fig:method}
\end{figure*}

\subsection{Residual Policy Learning}
Residual policy learning~\cite{ResiP, RPL, RRL_RC} decomposes control into a nominal action and a learned correction. By learning only the residual, the policy is constrained to local corrections, which improves exploration, sample efficiency, and training stability compared to end-to-end learning~\cite{RPL}.

For shared autonomy, residual formulations~\cite{DRL_SA} naturally separate roles: the pilot provides high-level motion commands, while the residual copilot applies low-level corrections for alignment, contact regulation, and robustness, which preserves human intent and improves performance.

We adopt this abstraction and train a residual copilot with model-free reinforcement learning to correct teleoperation commands from sparse task rewards, without requiring explicit intent modeling or expert priors.

\subsection{Real-to-Sim-to-Real Learning}
Real-to-sim-to-real pipelines use real-world data to construct task-relevant simulations for scalable and transferrable policy learning, followed by deployment back to the physical system. Prior work~\cite{robo_gs, ASAP, GS_bridge, RialTo,huang2025vtrefine,dexpoint, touch-dexterity} typically digitalizes geometry, dynamics, and sensing, often with domain randomization to improve sim-to-real robustness.

In shared autonomy, the simulator must also model the human pilot, since the copilot conditions on the pilot's actions. As a result, the quality of learned assistance depends on how faithfully human behavior is represented during training.

We therefore use real teleoperation data to instantiate a human surrogate in simulation, then train residual copilot policies entirely in simulation with model-free reinforcement learning, and deploy them directly in the real world to assist human operators. Beyond human modeling, the real-to-sim stage also digitalizes task and dynamics details, including environment recreation and system identification of controller and physics parameters, improving simulation fidelity for contact-rich interaction.

\section{Method}

\subsection{Problem Definition}
We model the shared autonomy problem as a partially observable Markov decision process (POMDP)~\cite{POMDP_Paper}:
\[
\mathcal{M}=\langle \mathcal{X},\, \mathcal{A},\, \mathcal{T},\, \mathcal{R},\, \Omega,\, O,\, \gamma\rangle.
\]
At each timestep, the full state $x_t \in \mathcal{X}$ is defined as $x_t=(s_t, g)$, where $s_t \in \mathcal{S}$ denotes the environment state and $g \in \mathcal{G}$ represents the human pilot's latent goal, which is unobservable to the copilot (assistive agent). The environment state $s_t$ includes the robot end-effector pose and velocity in task space, the gripper state, and the object poses expressed in the end-effector frame.

Observing the environment, the pilot provides a goal-implicit control command $a_t^h = [p, q, u] \in \mathcal{A}$ via teleoperation, consisting of a target task-space end-effector pose $[p, q]$ and a gripper command $u$. Rather than explicitly inferring the pilot's latent goal $g$, we treat intent as being expressed implicitly through teleoperation commands.

The copilot observes both the environment state and the pilot input, i.e., $o_t=(s_t, a_t^h) \in \Omega$. We parameterize the copilot as a residual policy, which we refer to as the \texttt{Residual Copilot}, treating the pilot command $a_t^h$ as a base action and learning a corrective delta. At each step, the copilot predicts a normalized residual action $a_t^{\mathrm{res}} \sim \pi_r(\cdot \mid o_t)$, and the final action command is
\[
a_t = a_t^h \oplus \alpha a_t^{\mathrm{res}},
\]
where $\oplus$ composes translational and gripper increments additively and applies rotational increments multiplicatively in task space (see Appendix~\ref{representation_space}). The residual scale $\alpha$ maps normalized residual action to controllable physical magnitudes. This action induces a state transition $x_{t+1} \sim \mathcal{T}(\cdot \mid x_t, a_t)$ and yields a reward $r_t=\mathcal{R}(x_t, a_t)$.

Our objective is to learn the \texttt{Residual Copilot} policy $\pi_r(a_t^{\mathrm{res}} \mid o_t)$ that maximizes the expected discounted return, without access to the environment dynamics $\mathcal{T}$, the pilot goal space $\mathcal{G}$, or the pilot policy $\pi_h(a_t^h \mid x_t)$. The \texttt{Residual Copilot} is parameterized as a Gaussian policy with an MLP backbone and optimized using model-free reinforcement learning with PPO~\cite{PPO}.

We decompose the reward as $\mathcal{R}=\mathcal{R}_{\text{general}}+\mathcal{R}_{\text{success}}$, where $\mathcal{R}_{\text{general}}$ captures goal-agnostic objectives shared across a set of tasks, and $\mathcal{R}_{\text{success}}$ provides a task-specific sparse success signal. In our fine-grained assembly experiments, $\mathcal{R}_{\text{general}}$ includes termination and contact-force penalties, as well as shaping terms for assembly-axis alignment, upright end-effector orientation, and action regularization (Table~\ref{tab:reward_terms}).

\subsection{Human Surrogate Model}
\label{human_surrogate}
Training the copilot directly with human pilots in the loop is costly and difficult to scale. Following prior work~\citep{DRL_SA, RPL_SA}, we train in simulation using a surrogate pilot in place of the human pilot.

A suitable human surrogate must satisfy two requirements: it must remain coherent under the distribution shift induced by copilot exploration (\emph{mutual dependency}); and it must be constructible from limited data, as otherwise the problem effectively reduces to training a fully autonomous policy from a large dataset.

We therefore adopt a non-parametric $k$-nearest-neighbor (kNN) surrogate $\pi_h^{\mathrm{kNN}}$ (the \texttt{kNN Pilot}) built from a small demonstration set that may include unsuccessful demonstrations. By retrieving actions directly from the empirical demonstration manifold, the \texttt{kNN Pilot} avoids extrapolation beyond distributional support; assuming the human pilot acts along a consistent action manifold with or without assistance, this inductive bias promotes stable behavior under moderate distribution shift while remaining inherently data-efficient.

For neighbor retrieval, we define a weighted distance over  translation $p$, rotation $q$, and gripper $u$ components using only proprioceptive end-effector commands:
\begin{equation}
\label{eq:action_distance}
d(a,a') = \alpha_1 \|p-p'\|_2 + \alpha_2 d_{\mathrm{SO}(3)}(q,q') + \alpha_3 |u-u'|,
\end{equation}
where $d_{\mathrm{SO}(3)}$ denotes quaternion geodesic distance and $\{\alpha_i\}$ are tuned by minimizing nearest-neighbor prediction error on the demonstration set.

To improve robustness during copilot training, we augment \texttt{kNN Pilot} with two mechanisms: (1) action chunking to preserve short-horizon structure and avoid myopic switching between neighbors, and (2) local stochastic perturbations to expand coverage under distribution shift.

\textbf{Action chunking.}
At runtime, $\pi_h^{\text{kNN}}:\mathcal{S}\!\rightarrow\!\mathcal{A}$ retrieves the $k$ nearest demonstrated commands under Eq.~\eqref{eq:action_distance} and samples one via a temperature-scaled softmax over negative distances. Instead of returning a single command, the surrogate outputs a short \emph{action chunk} of consecutive demonstrated commands from the selected neighbor, preserving temporal consistency.

\textbf{Smooth local perturbations.}
Under residual assistance, the copilot may drive the system outside the empirical support of the demonstrations, making pure retrieval brittle. To locally expand coverage, we inject i.i.d.\ local perturbations composed with the current command using the same operator~$\oplus$. At each step, we sample $\varepsilon_t \sim \mathcal{U}([l,h]^d)$ and form $\delta a_t = m_t\,\varepsilon_t$, where the gate $m_t \in [0,1]$ evolves as $m_t = \beta m_{t-1} + (1-\beta)b_t$ with $b_t \sim \mathrm{Bernoulli}(p_{\mathrm{on}})$. The perturbation is applied via $a_t^{h} \leftarrow a_t^{h} \oplus \delta a_t$. The Bernoulli parameter $p_{\mathrm{on}}$ controls activation frequency, while $\beta$ governs the smooth rise and decay of noisy phases.

\subsection{Admittance Control}
To safely complete contact-rich manipulation tasks, we execute commands with task-space compliance via admittance control, allowing the end-effector to yield under contact while tracking goal commands.
We model compliance with a virtual spring--mass--damper relationship~\cite{ACL, adm_ctrl} between task-space motion and the measured external wrench:
\begin{equation*}
\label{eq:admittance}
f \;=\; M\ddot{\xi} \;+\; D\dot{\xi} \;+\; K(\xi-\xi_{\mathrm{ref}}),
\end{equation*}
where $\xi\in\mathbb{R}^N$ denotes task-space pose coordinates, $f\in\mathbb{R}^N$ is the measured wrench, and $(M,D,K)$ are the virtual inertia, damping, and stiffness; $\xi_{\mathrm{ref}}$ is the reference pose at rest.

\subsection{Real-to-Sim-to-Real}
To improve human surrogate fidelity, reduce simulation bias, and learn reliable assistance behaviors, we adopt a real-to-sim-to-real pipeline. In the real-to-sim stage, we digitalize both the human surrogate and task environment. A small set of real teleoperation demonstrations is used to fit the \texttt{kNN Pilot} and to calibrate the simulator by tuning low-level control and physical parameters to match real trajectories. We optimize joint-space PID gains $(K_p, K_i, K_d)$ for the arm and gripper; task-space admittance gains, including stiffness $(K_x, K_r)$, damping $(D_x, D_r)$, and virtual mass/inertia $(M_x, M_r)$; and physical parameters affecting contact dynamics, including friction coefficient $\mu$ and object center of mass $m_{\mathrm{CoM}}$.

In the sim-to-real stage, we apply domain randomization over controller parameters and object pose observations to account for controller discrepancies and real-world pose estimation error (Appendix~\ref{app:dmr}).

\section{Experiments}
\begin{figure*}[t]
  \centering
  \includegraphics[width=\textwidth]{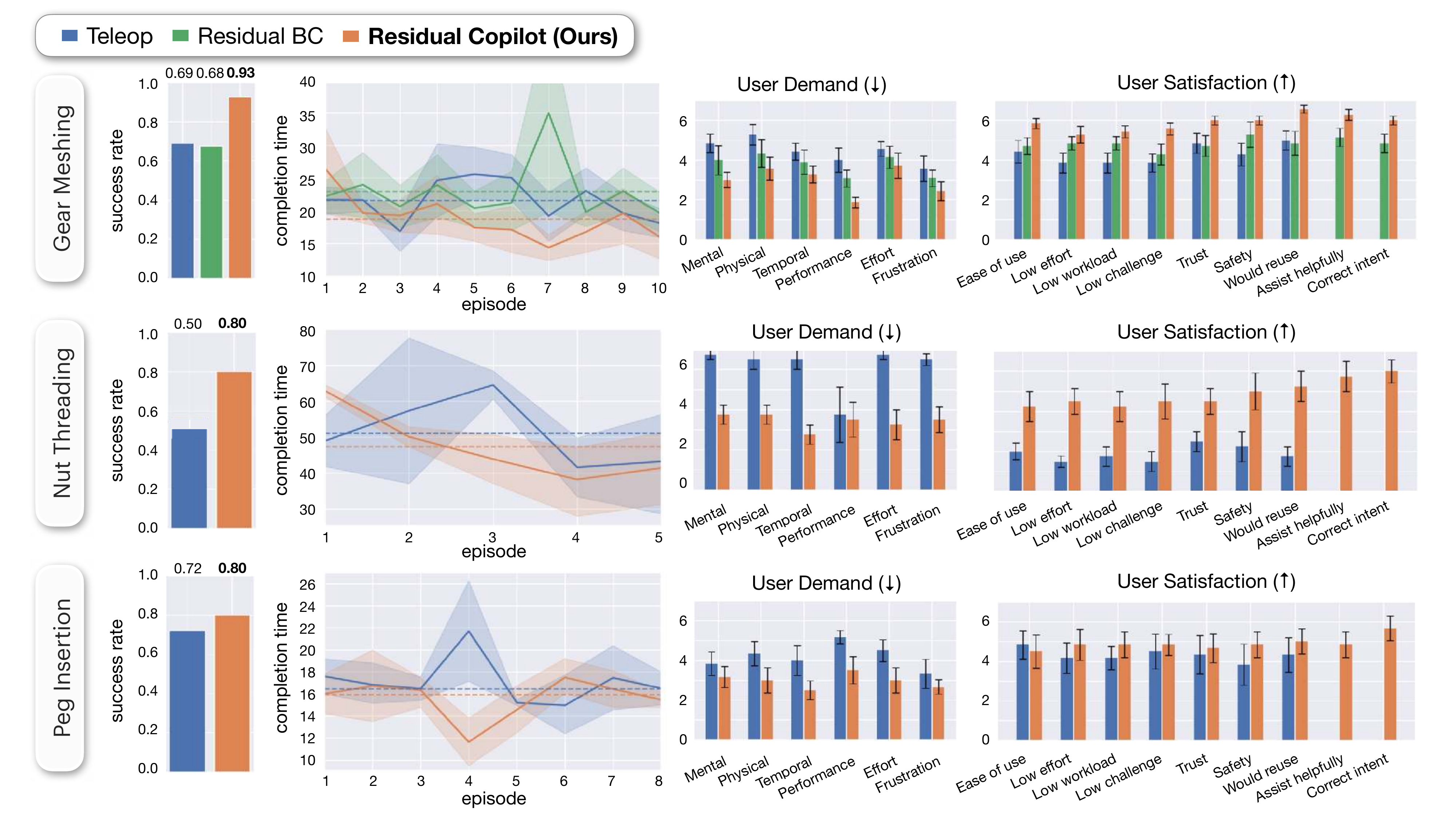}
  \caption{
    \small \textbf{Human experiment results across three contact-rich assembly tasks.}
    Rows correspond to Gear Meshing, Nut Threading, and Peg Insertion.
    \textbf{Left:} success rate and mean completion time across successful episodes (solid line; shaded region denotes standard deviation; dashed line indicates per-method mean).
    \textbf{Right:} subjective workload (NASA-TLX subscales; lower is better) and user satisfaction (higher is better).
    Across tasks, the \texttt{Residual Copilot} improves success rate and can effectively reduce completion time, while lowering reported workload and increasing user satisfaction compared to direct teleoperation and a residual baseline.}
  \label{fig:human_exp}
\end{figure*}

In this section, we evaluate our \texttt{Residual Copilot} in a suite of fine-grained, contact-rich manipulation tasks. We aim to address the following questions:
(\textbf{Q1}) Can our \texttt{Residual Copilot} improve a human operator's performance in fine-grained, contact-rich manipulation tasks? 
(\textbf{Q2}) Does our choice of human surrogate \texttt{kNN Pilot} outperform prior choices of human surrogates and guided diffusion baselines?
(\textbf{Q3}) Does data collected with the \texttt{Residual Copilot} lead to improved downstream imitation policy performance?

\begin{figure*}[t]
  \centering
  \includegraphics[width=1.\textwidth]{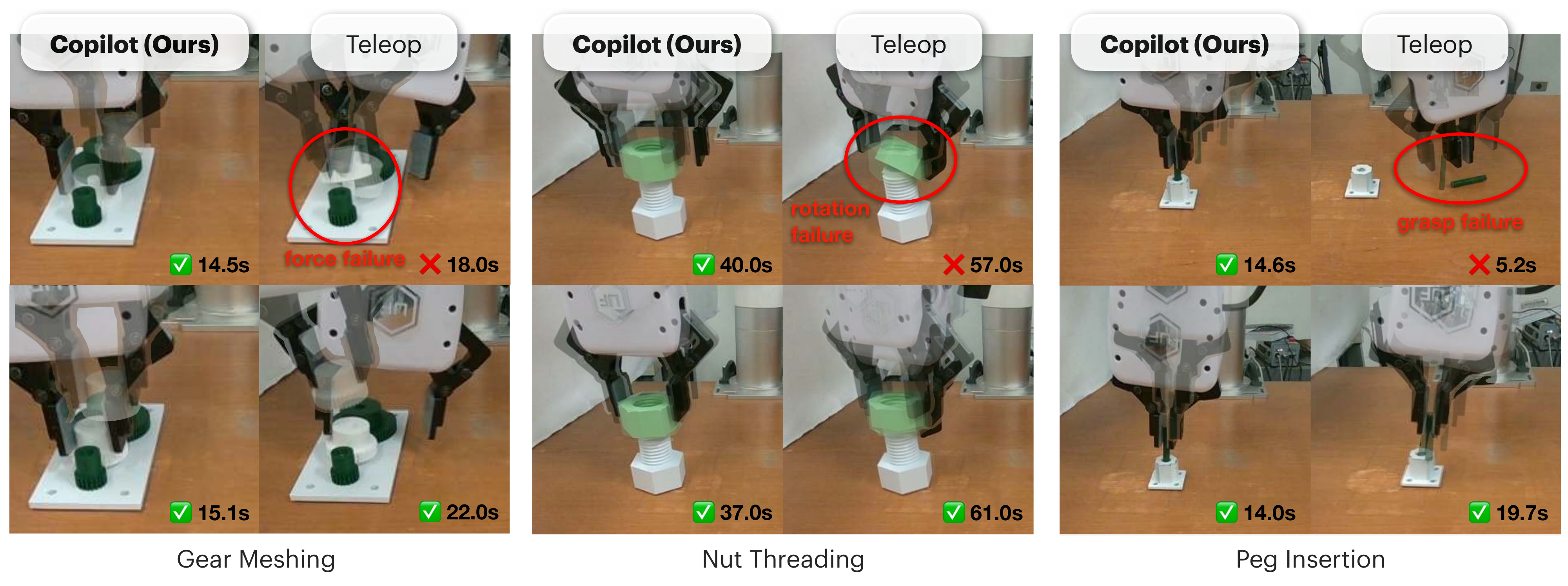}
  \vspace{-10pt}
  \caption{
    \small \textbf{User study qualitative    results.} Representative trajectory temporal overlays comparing unassisted teleoperation and our \texttt{Residual Copilot} on three high-precision tasks. \textbf{Top row (effectiveness):} our copilot enables reliable task completion in cases where teleoperation fails due to force, rotation, or grasp errors. \textbf{Bottom row (efficiency):} when both succeed, our copilot consistently reduces completion time and exemplifies smooth motions. Together, these results demonstrate that our method provides both reliable and efficient assistance in challenging manipulation scenarios.}
    \label{fig:qual}
\end{figure*}

\textbf{Experiment Setup.}
To address these questions, we evaluate the 
\texttt{Residual Copilot} through a real-world user study across three 
contact-rich assembly tasks (Sec.~\ref{exp1}), a real-world and simulation 
analysis of pilot model choice and robustness (Sec.~\ref{exp2}), and a 
downstream imitation learning comparison (Sec.~\ref{exp3}).

We consider three representative fine-grained, contact-rich assembly tasks from the NIST board \#1~\cite{nist_assembly_2022} (radial clearance $< 1\,\mathrm{mm}$):
\begin{itemize}
    \item \textbf{Gear Meshing:} pick up the medium gear, insert it onto its shaft, and mesh with the other gears. A trial succeeds only if the gear remains grasped until insertion completes.
    \item \textbf{Nut Threading:} pick up an M32 nut and tighten it onto the bolt by at least $180^\circ$. Due to hardware limits of GELLO, participants perform three $60^\circ$ turns.
    \item \textbf{Peg Insertion:} pick up an $8\,\mathrm{mm}$-long peg and insert it into the base. A trial succeeds only if the peg reaches the bottom of the receptacle.
\end{itemize}
For all tasks, failure is defined as dropping the object prior to success or exceeding the xArm safety force threshold.
To ensure a fair comparison, we fix the robot to a common initial pose and randomize the initial object placement within the workspace at the start of each trial.

In the real-world setup, participants teleoperate a UFactory xArm7 using GELLO~\cite{gello} in 7-DoF task space (end-effector pose and gripper command). We attach a wrist-mounted force-torque sensor to implement admittance control. We use an external Intel RealSense D455 camera and estimate object poses with FoundationPose~\cite{foundation_pose}.
In simulation, we use the NVIDIA Isaac Lab~\cite{isaaclab} gear-meshing and peg-insertion environments from Factory~\cite{factory}, and implement the M32 nut-and-bolt task similarly with SDF-based collision. We read ground-truth contact forces at the end-effector for admittance control.

\textbf{Subject Allocation.}
We recruited 12 novice teleoperators (fewer than 20 prior teleoperation 
trajectories) for gear meshing and peg insertion, and 4 experienced teleoperators (at least 50 prior teleoperation trajectories) for nut threading, totaling over 20 hours of teleoperation trials across the three tasks.
Participants received task objectives and a high-level description of each method (see Appendix for details).
Each participant completed a 5-minute warm-up to familiarize with the interface and system dynamics.
Following prior work~\cite{DRL_SA, RPL_SA}, we interleave direct teleoperation and copilot-assisted teleoperation to control for learning effects.
After completing all trials, participants filled out NASA-TLX and user-satisfaction questionnaires.

\textbf{Baselines.}
We implement the following pilot baselines to isolate the effect of pilot modeling on copilot learning and to systematically evaluate robustness under varying assumptions about human behavior:
\begin{itemize}
    \item \texttt{BC Pilot}: Following~\citet{RPL_SA}, we learn a behavioral surrogate by imitation. For each task, we collect fewer than $5$ minutes of real teleoperation data, apply geometric data augmentation, and train a Diffusion Policy (DP)~\cite{dp} to model the pilot's action distribution.

    \item \texttt{Expert-based Pilots}: Following~\citet{DRL_SA}, we construct a near-optimal \texttt{Expert Pilot} by training a DP on 2000 successful rollouts (200--400k transitions) generated by our \texttt{Residual Copilot}. We additionally instantiate a \texttt{Laggy Pilot} (previous action repeated with probability $0.8$) and a \texttt{Noisy Pilot} (smooth gated noise injected with probability $0.5$; see Section~\ref{human_surrogate}), following the abstract pilot perturbations proposed in~\citet{DRL_SA} to model delayed reaction and execution errors, respectively.
\end{itemize}

Using these pilots, we train the following copilot baselines:

\begin{itemize}
    \item \texttt{GD Copilots}: Following prior guided diffusion (GD) shared autonomy formulations~\cite{DexGen}, we train a diffusion copilot that denoises actions conditioned on pilot commands. We consider two variants differing in training data optimality: \texttt{GD BC}, trained on the augmented teleoperation dataset used for the \texttt{BC Pilot}; and \texttt{GD Expert}, trained on the expert rollouts used for the \texttt{Expert Pilot} (an impractical upper-bound baseline included to characterize the limits of GD methods).

    \item \texttt{Residual BC}: A residual RL copilot trained to assist the \texttt{BC Pilot} using the same RL setup as our method, differing only in the choice of pilot model.
\end{itemize}

\begin{table*}[t]
\centering
\renewcommand{\arraystretch}{1.05}
\scriptsize
\setlength{\tabcolsep}{4pt}
\begin{tabular}{@{}l|ccc|ccc|ccc|ccc|ccc@{}}
\toprule 
\multirow{3}{*}{\textbf{Copilot}} & \multicolumn{15}{c}{\textbf{Eval. Pilot}} \\ \cmidrule{2-16} 
& \multicolumn{3}{c|}{\texttt{Laggy Pilot}} 
& \multicolumn{3}{c|}{\texttt{Noisy Pilot}} 
& \multicolumn{3}{c|}{\texttt{Expert Pilot}} 
& \multicolumn{3}{c|}{\texttt{BC Pilot}} 
& \multicolumn{3}{c}{\texttt{kNN Pilot}} \\ 
& \textbf{Gear} & \textbf{Peg} & \textbf{Nut} 
& \textbf{Gear} & \textbf{Peg} & \textbf{Nut} 
& \textbf{Gear} & \textbf{Peg} & \textbf{Nut} 
& \textbf{Gear} & \textbf{Peg} & \textbf{Nut} 
& \textbf{Gear} & \textbf{Peg} & \textbf{Nut} \\ 
\midrule
\texttt{No Copilot}
& \underline{0.97} & \textbf{0.94} & \underline{0.49}
& 0.92 & \underline{0.91} & 0.24
& \textbf{0.99} & \textbf{0.99} & \underline{0.61}
& 0.84 & 0.71 & 0.03
& 0.87 & 0.85 & 0.16 \\
\texttt{GD Expert}
& 0.96 & 0.89 & 0.24
& \underline{0.95} & \underline{0.91} & \underline{0.25}
& \colorbox{hlgray}{0.94} & \colorbox{hlgray}{0.90} & \colorbox{hlgray}{0.28}
& \underline{0.94} & \textbf{0.89} & \textbf{0.34}
& \underline{0.95} & \underline{0.90} & \underline{0.27} \\
\texttt{GD BC}
& 0.63 & 0.56 & 0.00
& 0.65 & 0.56 & 0.00
& 0.62 & 0.53 & 0.01
& \colorbox{hlgray}{0.63} & \colorbox{hlgray}{0.52} & \colorbox{hlgray}{0.01}
& 0.66 & 0.55 & 0.00 \\
\texttt{Residual BC}
& 0.96 & 0.44 & 0.00
& 0.91 & 0.44 & 0.00
& \underline{0.98} & 0.44 & 0.00
& \colorbox{hlgray}{0.70} & \colorbox{hlgray}{0.40} & \colorbox{hlgray}{0.00}
& 0.86 & 0.46 & 0.02 \\
\textbf{\texttt{Residual Copilot (Ours)}}
& \textbf{0.99} & \underline{0.92} & \textbf{0.65}
& \textbf{0.96} & \textbf{0.92} & \textbf{0.69}
& \underline{0.98} & \underline{0.93} & \textbf{0.74}
& \textbf{0.97} & \underline{0.83} & \underline{0.21}
& \colorbox{hlblue}{\textbf{1.00}} & \colorbox{hlblue}{\textbf{0.99}} & \colorbox{hlblue}{\textbf{0.81}} \\
\bottomrule
\end{tabular}%

\vspace{3pt}
\caption{
    \small \textbf{Cross-pilot generalization and robustness in simulation.}
    Rows denote copilots trained with different training pilots; columns evaluate each trained copilot under distinct \emph{evaluation} pilots across three tasks. Each cell reports task progression (over 1000 simulation trials); see Appendix for metric details.
    Gray and blue diagonal entries indicate in-distribution evaluation, i.e., a copilot evaluated with the pilot it was trained on; off-diagonal entries measure generalization to unseen pilots.
    The \texttt{Residual Copilot} maintains consistently high performance across evaluation pilots, demonstrating robustness to pilot mismatch and perturbations such as lag and action noise.}
\label{tab:sr_merged}
\end{table*}

\subsection{Copilot Performance}
\label{exp1}
To address \textbf{Q1}, we conduct the user study to compare direct and copilot-assisted teleoperation, evaluating both objective task performance and subjective user experience across three fine-grained, contact-rich assembly tasks.
Fig.~\ref{fig:human_exp} summarizes success rate and completion time, along with NASA-TLX workload scores and user satisfaction.
Overall, residual assistance consistently outperforms direct teleoperation, yielding higher success rates, faster execution, and improved subjective experience across tasks.

The benefits of residual assistance manifest differently depending on task structure and failure modes.
For \textbf{Nut Threading}, which requires sustained axis-constrained rotation and precise orientation control under contact, residual rotational stabilization and acceleration are critical for maintaining thread engagement.
Without assistance, operators frequently lose alignment or apply insufficient rotational motion, leading to premature disengagement, as shown in Fig.~\ref{fig:qual}.
The copilot mitigates these issues by stabilizing orientation and amplifying effective rotational commands, improving success by up to $30\%$ while reducing the need for repeated corrective motions.
For \textbf{Gear Meshing}, success depends on achieving precise insertion depth followed by small corrective rotations to align gear teeth.
Direct teleoperation often fails due to slight misalignment that is difficult for operators to perceive or correct through the interface.
The \texttt{Residual Copilot} learns to reliably complete insertion and apply subtle rotational corrections at contact, reducing the operator’s low-level alignment burden while preserving their high-level intent.
As a result, operators achieve higher success with fewer failed insertion attempts.
For \textbf{Peg Insertion}, failures arise not only during insertion but also 
from unreliable grasping.
In this setting, the copilot stabilizes the end-effector at an effective pre-grasp pose, centered above the peg, while leaving the timing of grasp initiation entirely to the user.
This division of roles preserves operator control over task sequencing while improving grasp reliability, leading to higher overall success rates.

Completion-time trends further verify these improvements.
Across tasks, the \texttt{Residual Copilot} generally reduces completion time by decreasing the number of failed attempts, corrective motions, and recovery behaviors required to complete each trial.
These objective gains align with subjective reports (Fig.~\ref{fig:human_exp}, right), where participants report lower workload across multiple NASA-TLX subscales and higher overall satisfaction.
Together, these results indicate that residual assistance not only improves task outcomes, but also reduces cognitive and physical burden while maintaining a clear and intuitive division of control between the human and the copilot.

\subsection{Human Surrogate Analysis}
\label{exp2}
To address \textbf{Q2}, we compare our \texttt{Residual Copilot} against \texttt{Residual BC} with novice teleoperators on gear meshing (Fig.~\ref{fig:human_exp}, top row). Our copilot improves objective performance (success rate and completion time) and receives higher subjective ratings. In particular, participants report higher \emph{correct intent}, suggesting that training with the \texttt{kNN Pilot} promotes more stable exploration and better pilot generalization, yielding more reliable local corrections that preserve human intent.

We further evaluate surrogate quality in simulation by training copilots under different pilot models and testing them across evaluation pilots (Table~\ref{tab:sr_merged}). Across tasks and evaluation pilots, our \texttt{Residual Copilot} achieves the highest in-distribution performance and exhibits the smallest degradation under pilot mismatch, indicating improved robustness to distribution shift.

Under suboptimal pilots, \texttt{Residual BC} achieves consistently lower task progression, suggesting that it fails to explore success-relevant regions when distribution shifts arise from copilot-induced exploration under sparse rewards, leading to premature convergence to local minima where only generalized rewards are obtained.

Finally, \texttt{GD Copilots} are highly sensitive to the pilot model. In particular, guiding \texttt{GD Expert} with the \texttt{Expert Pilot} distribution reduces performance relative to \texttt{No Copilot} under the \texttt{Expert Pilot}. For contact-sensitive tasks such as Nut Threading, which require consistent axis-constrained rotation and precise contact regulation, \texttt{GD Expert} does not exceed 34\% progression under any pilot model, despite achieving 61\% progression autonomously. This indicates that test-time guidance can steer actions away from task-optimal regions when the pilot distribution is suboptimal. Conversely, steering \texttt{GD BC} with stronger priors does not yield meaningful improvements. Overall, this asymmetry suggests that test-time guidance alone is insufficient to reconcile action optimality with preservation of human intent.

\subsection{Data Quality Comparison}
\label{exp3}

To address \textbf{Q3}, we compare the quality of demonstrations collected under \texttt{Residual Copilot} assistance versus direct teleoperation for downstream imitation learning.
We train a vision-based Diffusion Policy (DP) under two conditions:
\begin{enumerate}
    \item[(i)] \textbf{Matched attempts.} We randomly sample five participants from the gear-meshing study and train a DP using all successful demonstrations collected from an equal number of teleoperation attempts for each method.
    \item[(ii)] \textbf{Matched successes.} We randomly sample five participants, compute the minimum number of successful trials across the two methods, and train a DP using the same number of successful demonstrations for both.
\end{enumerate}

Under \emph{matched attempts}, where both methods contribute the same number of teleoperation trials, the DP trained on copilot-assisted data achieves substantially higher task progression than the DP trained on teleoperation data.
As shown in Table~\ref{tab:data_quality}, the DP trained on copilot-collected data reaches 18/20 grasp successes and 11/20 insertion successes, compared to 7/20 grasps and 1/20 insertion for teleoperation.
This gap indicates that copilot-assisted demonstrations provide more learnable supervision for the downstream DP under the same data-collection attempt budget.

\begin{table}[t]
  \centering
  \begin{tabular}{l cc cc}
    \toprule
    & \multicolumn{2}{c}{\textbf{Matched attempts}} & \multicolumn{2}{c}{\textbf{Matched successes}} \\
    \cmidrule(lr){2-3} \cmidrule(lr){4-5}
    \textbf{Method} & \textbf{Grasp} & \textbf{Insert} & \textbf{Grasp} & \textbf{Insert} \\
    \midrule
    Teleop & 7/20 & 1/20 & 6/20 & 0/20 \\
    \textbf{Residual (Ours)} & \textbf{18/20} & \textbf{11/20} & \textbf{19/20} & \textbf{9/20} \\
    \bottomrule
  \end{tabular}
    \caption{\small \textbf{Downstream policy progression from copilot vs.\ teleoperation data (Gear Meshing).}
  Diffusion policies are trained on demonstrations collected under \texttt{Residual Copilot} assistance or direct teleoperation, using matched attempts and matched successes to control for data budget. Results show copilot assistance improves learnability and consistency of the collected demonstrations.}
  \label{tab:data_quality}
\end{table}

More importantly, under \emph{matched successes}, this advantage persists even when both datasets contain the same number of successful demonstrations.
The DP trained on copilot-assisted data achieves 19/20 grasp success and 9/20 insertion success, whereas the DP trained on teleoperation data achieves 6/20 grasps and 0/20 insertions.
Because the number of successful demonstrations and the model capacity are controlled, this comparison isolates demonstration quality rather than quantity.

We attribute this difference to qualitative properties of the collected trajectories, illustrated in Fig.~\ref{fig:qual}. Residual copilot assistance produces demonstrations that are more consistent in alignment, contact timing, and approach strategy, while still reflecting human intent.
In contrast, successful teleoperation demonstrations often include larger corrective motions, abrupt recoveries, or incidental contacts, increasing trajectory variability.
Such variability makes the action distribution harder for the diffusion policy to model and degrades generalization.

Overall, these results indicate that \texttt{Residual Copilot} assistance improves the structure and consistency of successful demonstrations.
This leads to downstream policies that generalize more reliably, even when trained on the same number of successes, highlighting the value of shared autonomy as a tool for scalable, high-quality demonstration collection.

\section{Conclusion} 
\label{sec:conclusion}
We presented a real-to-sim-to-real shared autonomy framework for fine-grained, contact-rich teleoperation.
Our approach trains a residual copilot using model-free reinforcement learning in simulation, driven by a lightweight kNN human surrogate fit from fewer than five minutes of real teleoperation data.
By learning only low-level corrective behaviors on top of human commands, the copilot preserves operator intent while avoiding reliance on brittle parametric behavioral-cloning pilots or expert autonomous priors.

Across both simulation and a real-world user study on nut threading, gear meshing, and peg insertion, our system improves task success for novice operators and execution efficiency for experienced operators relative to direct teleoperation and prior shared-autonomy baselines. Beyond execution performance, copilot-assisted teleoperation produces demonstrations more efficiently and with greater consistency and structure, leading to improved downstream imitation learning policy performance.

A key strength of the proposed approach is its residual formulation, learned through a real-to-sim-to-real pipeline.
Because the copilot operates as a correction on top of user commands, it is largely agnostic to teleoperator skill level and can be integrated with existing teleoperation interfaces without modifying user control strategies.
This makes shared autonomy a practical tool not only for assistive execution, but also for scalable, high-quality data collection in fine-grained, contact-rich manipulation.

One limitation of our work is the assumption that user behavior during data collection matches behavior under assistance.
In practice, users may adapt and exhibit emergent strategies once a copilot is present, introducing additional distribution shift.
An important direction for future work is to explicitly model this co-adaptation, for example by updating the human surrogate online or learning interaction policies that jointly account for human behavior and copilot assistance.

\section*{Acknowledgment}
This work was partially supported by the DARPA TIAMAT program (HR0011-24-9-0430), the Toyota Research Institute, NSF Award \#2409661, Samsung Research America, and an Amazon Research Award (Fall 2024). This article solely reflects the opinions and conclusions of its authors and should not be interpreted as necessarily representing the official policies, either expressed or implied, of the sponsors.

We would like to thank Kaifeng Zhang, Hanxiao Jiang, Fangyu Wu, Shivansh Patel, Jaisel Singh, and all other members of the RoboPIL Lab for helpful discussions during the project. 
We would also like to thank each participant of the user study for their time and feedback.

\bibliographystyle{plainnat}
\bibliography{references}

\clearpage
\begin{center}
\textsc{\Large Appendix}
\end{center}

\newcommand{\DoToC}{
    \hypersetup{linkcolor=black}
  \startcontents
  \printcontents{}{1}{\textbf{Contents}\vskip3pt\hrule\vskip3pt}
  \vskip7pt\hrule\vskip3pt
}

\appendices

\DoToC
\vspace{5pt}

\section{Additional Technical Details}
\subsection{Hardware System}
\subsubsection{Robot Setup}
\begin{figure}[t]
    \centering
    \includegraphics[width=\linewidth]{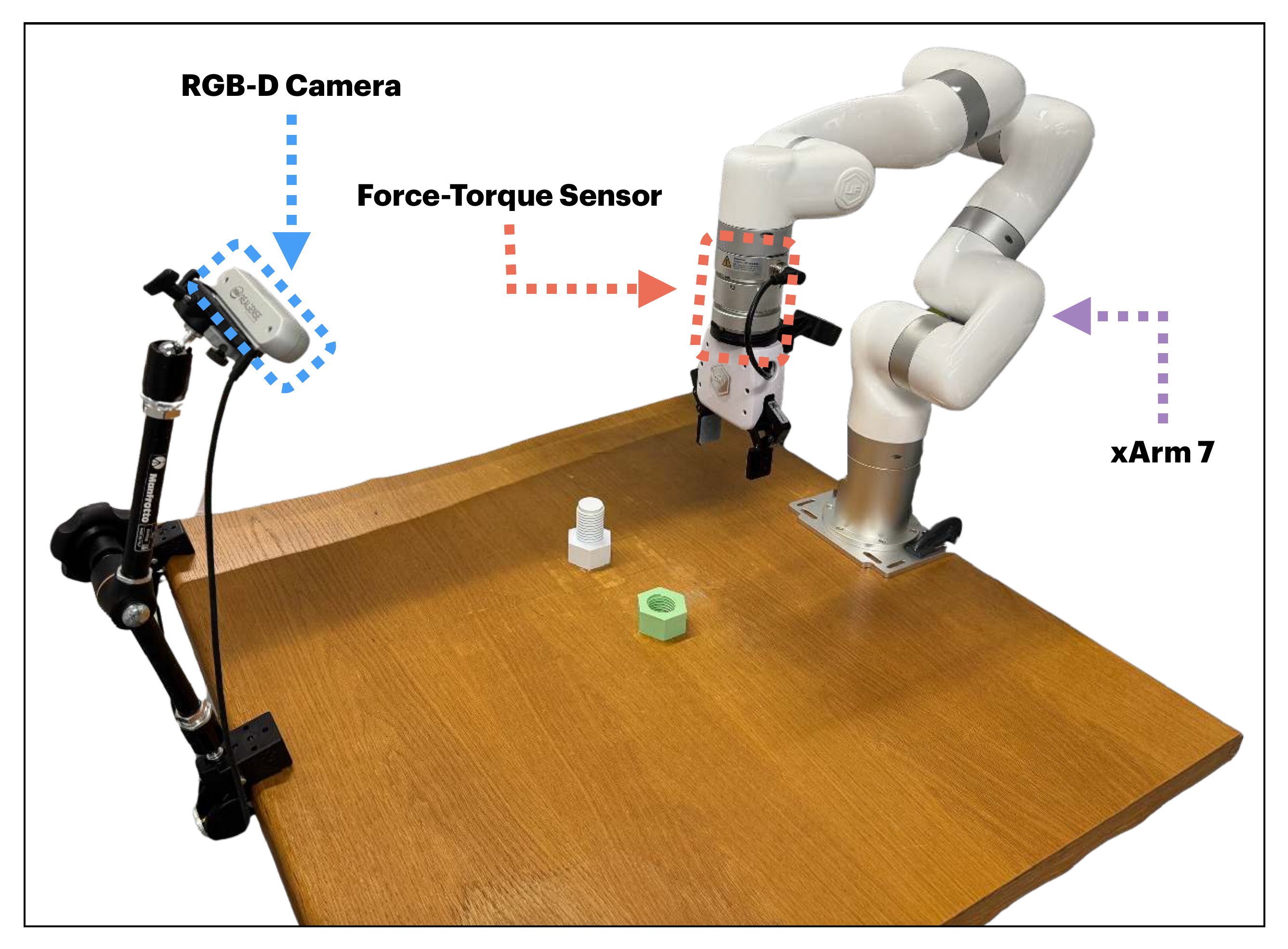}
    \caption{\small \textbf{xArm 7 Setup.}}
    \label{fig:setup}
\end{figure}

We use a UFactory xArm 7 mounted on a tabletop, shown in Fig.~\ref{fig:setup}. The robot has 7 degrees of freedom. A wrist-mounted force--torque sensor is attached at the end-effector (link 7), with a standard xArm parallel gripper mounted to the sensor. An Intel RealSense D455 RGB-D camera is placed in front of the robot to provide an overhead view of the workspace.

\subsubsection{Residual Implementation}
We use GELLO~\cite{gello} as the teacher arm for teleoperation.
For direct teleoperation, GELLO joint angles are streamed as joint-position commands directly to the xArm at 15\,Hz.
For residual-assisted teleoperation, we convert GELLO joints to a task-space end-effector pose via forward kinematics and treat it as a Cartesian target.

We additionally condition the residual copilot on the current observation and this Cartesian target; it predicts a corrective residual that is composed with the target before solving inverse kinematics to obtain robot joint targets, which are then streamed as joint-position commands to the xArm at 15\,Hz.
For safety, both methods clip the step-wise change in joint targets with an $\ell_2$ bound consistent with a joint-velocity limit.

\subsubsection{Admittance Control}
We enable the xArm 7’s built-in FT-sensor admittance mode at the end-effector, with 6-DoF compliance enabled (i.e., compliance along all translational and rotational axes) in the robot base frame. Damping is set to zero and we rely on the robot’s internal low-level servo for stability. Admittance control parameters can be found in Table \ref{tab:controller_params}.

\begin{table}[t]
\centering
\begin{tabular}{l l c c l}
\hline
Controller & Parameter & Real & Sim & Unit \\
\hline
\multirow{7}{*}{\makecell[l]{Task-space\\admittance\\controller}}
& $K_x$        & $1000$ & $200$   & N/m \\
& $K_r$        & $10$   & $100$   & Nm/rad \\
& $M_x$        & $0.1$  & $0.125$ & kg \\
& $M_r$        & $1\times10^{-3}$ & $0.015$ & kg$\cdot$m$^2$ \\
& $D_x$        & $0$    & $5$     & N$\cdot$s/m \\
& $D_r$        & $0$    & $1.2$   & N$\cdot$m$\cdot$s/rad \\
& adm. axes    & $6$    & $6$     & DoF \\
\hline
\multirow{3}{*}{\makecell[l]{Joint-space\\PID controller}}
& $K_p$        & --     & $200$   & N$\cdot$m/rad \\
& $K_i$        & --     & $20$    & N$\cdot$m/(rad$\cdot$s) \\
& $K_d$        & --     & $0$     & N$\cdot$m$\cdot$s/rad \\
\hline
\end{tabular}
\caption{\small \textbf{Controller parameters.} Real-world and simulation settings for the task-space admittance and joint-space PID controllers.}
\label{tab:controller_params}
\end{table}

\subsubsection{State Estimation}
We estimate object states using FoundationPose~\cite{foundation_pose} in a separate thread running at 15\,Hz. The environment asynchronously retrieves the latest object pose estimates at its own timestep. To improve sim-to-real alignment, we estimate the robot’s linear and angular velocities in both simulation and the real world using finite differences followed by low-pass filtering.

\begin{table}[t]
\centering
\begin{tabular}{l l l}
\hline
Group & Symbol(s) & Distribution / value \\
\hline
Ctrl. & $K_x,K_r,M_x,M_r$ & $\mathcal{U}(0.95,\,1.05)\odot(\cdot)$ \\
Pose  & $\Delta p_{\mathrm{fix}},\Delta p_{\mathrm{held}}$ & $\mathcal{N}(0,\,0.002^2 I)$ \\
Init. & $\Delta p_0,\Delta \theta_0$ & $\mathcal{N}(0,\,0.02^2 I)$,\; $\mathcal{N}(0,\,0.035^2 I)$ \\
Base & $L$ & $L \sim \mathcal{U}\{5,\dots,15\}$ \\
 & $\varepsilon_t$ & $\mathcal{U}([-0.6,0.6]^d)$ \\
 & $\beta,p_{\mathrm{on}}$ & $\beta=0.8,\; p_{\mathrm{on}}=0.5$ \\
\hline
\end{tabular}
\caption{\small\textbf{Domain randomization (DMR).} Per-episode sampling unless noted; base-action DMR is sampled per timestep. 
For base-action randomization, $L$ denotes the length of kNN-retrieved action chunks, $\varepsilon_t$ is i.i.d.\ residual noise, and $(\beta,p_{\mathrm{on}})$ parameterize a smooth stochastic gate controlling the activation frequency and temporal correlation of the noise.}
\label{tab:dmr}
\end{table}

\subsection{Residual Policy Training}
\subsubsection{Representation Spaces} 
\label{representation_space}
We define the following representations for the human surrogate (base) and residual action and observation spaces.

\textbf{Base action space.} We use an 8D action comprising an end-effector target pose in the robot base frame and a gripper command. The first three dimensions specify the target fingertip position, the next four specify the target fingertip orientation (quaternion), and the final dimension is a continuous gripper openness command in $[-1,1]$.

\textbf{Residual action space.} We use a 7D residual that is applied relative to the base action. The first three dimensions are a fingertip position delta, the next three are an axis--angle rotation delta, and the final dimension is a gripper openness delta.

\textbf{Residual observation space.} The residual policy receives a 35D vector consisting of fingertip position (3), fingertip quaternion (4), gripper openness (1), fingertip position relative to the fixed object (3), fingertip position relative to the held object (3), end-effector linear velocity (3), end-effector angular velocity (3), base action (8), and previous residual action (7).

\textbf{Residual composition.}
Let the base orientation be represented as a quaternion $q$ and the rotational residual as an axis--angle vector $\delta\theta \in \mathbb{R}^3$. We convert $\delta\theta$ to a quaternion $\delta q$ and update the orientation via $q' = \delta q \otimes q$, where $\otimes$ denotes quaternion multiplication. Translational and gripper residuals are applied additively.

\textbf{Normalization.} All action and observation dimensions are normalized to $[-1,1]$ using running mean and standard deviation statistics updated online during training and frozen during inference.

\subsubsection{Reward Terms}
\begin{table}[t]
\centering
\small
\setlength{\tabcolsep}{3pt}
\begin{tabularx}{\linewidth}{l l Y c}
\toprule
Category & Name & Term & Scale \\
\midrule
\multirow{5}{*}{$\mathcal{R}_{\text{general}}$}
& Regularization  & $-\lVert a_{\text{res}}\rVert$                      & $0.1$ \\
& Tilt Penalty    & $-\theta_{\text{tilt}}$                             & $1.0$ \\
& Force Penalty   & $-\phi(F)$                                           & $0.2$ \\
& Axis Align      & $\mathbf{1}[\lVert p_{xy}-p^*_{xy}\rVert < \epsilon]$ & $0.05$ \\
& Smoothness    & $-\lVert a_{\text{res}}^{prev} - a_{\text{res}} \rVert$ & 0.1 \\
& Termination     & $-\mathbf{1}[\text{fail / timeout}]$                 & $50.0$ \\
\midrule
$\mathcal{R}_{\text{success}}$
& Success         & $\mathbf{1}[\text{first success}]$                  & $30.0$ \\
\bottomrule
\end{tabularx}
\caption{\small \textbf{Reward terms.}
$a_{\text{res}}$ and $a_{\text{res}}^{prev}$ denote current and previous normalized residual actions, $\theta_{\text{tilt}}$ the end-effector tilt angle from upright,
$\phi(F)$ a clipped contact-force penalty, and $p_{xy}$ the held-object planar position.
The success reward is issued once per episode upon first satisfying the task-specific success condition. The termination condition is dropping the held object early on.}
\label{tab:reward_terms}
\end{table}

Table~\ref{tab:reward_terms} lists the complete reward terms. For Gear Meshing and Peg Insertion, we use a sparse success reward that triggers when the held object reaches a predefined relative pose corresponding to a successful insertion (position error $<2.5$\,mm). For \textsc{Nut Threading}, success requires both reaching the predefined assembly pose and achieving a cumulative $180^\circ$ rotation of the nut about its local yaw axis in the tightening direction while maintaining the prescribed screwing pose.

\subsubsection{Domain Randomization}
\label{app:dmr}
To improve robustness and sim-to-real transfer, we apply domain randomization (DMR) at multiple levels, summarized in Table~\ref{tab:dmr}. We use controller randomization to account for residual controller mismatch between simulation and hardware, object pose randomization to model real-world pose estimation errors, and initial configuration randomization to promote generalization. In addition, we randomize the human surrogate by injecting temporally correlated residual noise to improve robustness of both the surrogate and the residual copilot.

\begin{table}[t]
\centering
\renewcommand{\arraystretch}{1.08}
\setlength{\tabcolsep}{7pt}
\begin{tabular}{p{0.42\linewidth} p{0.42\linewidth}}
\hline
\textbf{Hyperparameter} & \textbf{Value} \\
\hline
\multicolumn{2}{l}{\textbf{Architecture}} \\
Backbone & LSTM (2 layers, 1024 units) \\
MLP & [512, 128, 64], ELU \\
Actor--critic & shared trunk \\
Policy head & Gaussian (learnable $\log\sigma$) \\
\hline
\multicolumn{2}{l}{\textbf{PPO / Optimization}} \\
$\gamma$ / $\lambda$ & 0.995 / 0.95 \\
LR schedule & adaptive (base lr $10^{-4}$) \\
KL threshold & 0.008 \\
Clip $\epsilon$ & 0.2 \\
Entropy coef. & 0.0 \\
Critic coef. & 2.0 \\
Grad clip & 1.0 \\
Update epochs & 4 \\
Rollout horizon / actors & 128 / 128 \\
Minibatch size & 512 \\
\hline
\end{tabular}
\caption{\small \textbf{RL hyperparameters.} Actor--critic architecture and PPO settings used for training.}
\label{tab:rl_hparams}
\end{table}

\begin{table}[t]
\centering
\renewcommand{\arraystretch}{1.08}
\setlength{\tabcolsep}{6pt}
\begin{tabular}{l c c}
\hline
\textbf{Category} & \textbf{State DP} & \textbf{Vision DP} \\
\hline
\multicolumn{3}{l}{\emph{Inputs / Outputs}} \\
Observation & $s_t \in \mathbb{R}^{20}$ & RGB $(200{\times}200)$ + $s_t \in \mathbb{R}^8$ \\
Action & $a_t \in \mathbb{R}^8$ & $a_t \in \mathbb{R}^8$ \\
Horizon ($H$) & 16 & 16 \\
Action steps & 8 & 8 \\
\hline
\multicolumn{3}{l}{\emph{Diffusion model}} \\
Noise scheduler & DDPM & DDPM \\
Train timesteps & 100 & 100 \\
Inference steps & 10 & 10 \\
$\beta$ schedule & Squared cosine & Squared cosine \\
Prediction target & $\epsilon$ (noise) & $\epsilon$ (noise) \\
FiLM modulation & Enabled & Enabled \\
\hline
\multicolumn{3}{l}{\emph{Architecture}} \\
Encoder & None & ResNet-18 \\
U-Net dims & [512, 1024, 2048] & [512, 1024, 2048] \\
Group norm groups & 8 & 8 \\
\hline
\multicolumn{3}{l}{\emph{Optimization}} \\
Batch size & 256 & 512 \\
Optimizer & Adam ($\mathrm{lr}=10^{-4}$) & Adam ($\mathrm{lr}=10^{-4}$) \\
LR schedule & Cosine & Cosine \\
Weight decay & $10^{-6}$ & $10^{-6}$ \\
Training steps & 40k & 10k \\
\hline
\end{tabular}
\caption{\small\textbf{Diffusion policy hyperparameters.} State- and vision-based variants share the diffusion process and training setup, differing only in input modality, encoder, and training scale.}
\label{tab:dp_hparams}
\end{table}

\begin{figure*}[t]
    \centering
    \includegraphics[width=\linewidth]{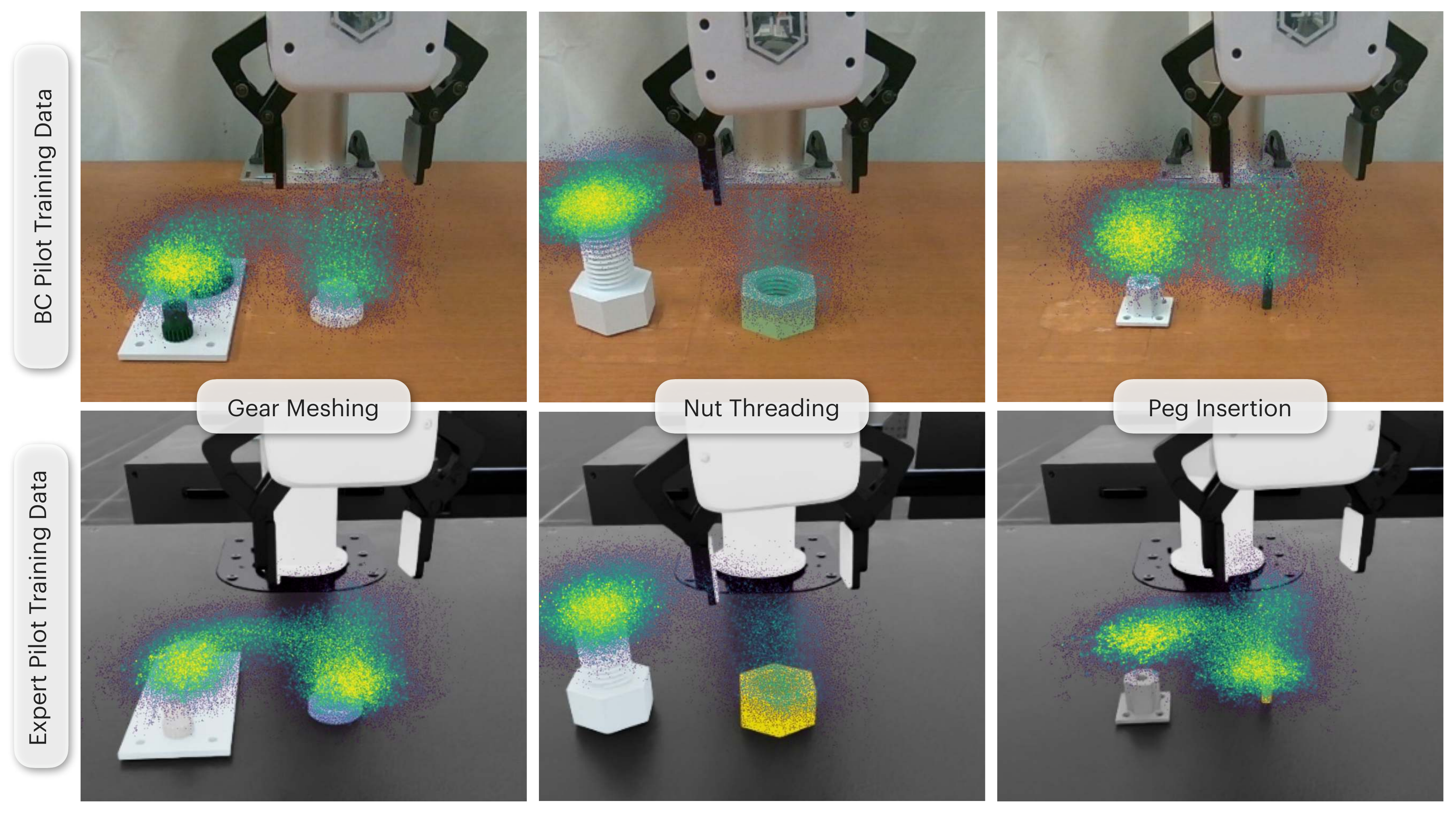}
    \caption{\small \textbf{State-based DP training data distributions.}
    Spatial coverage of 3D fingertip positions, downsampled to 100k points for visualization. The top row shows training data for \texttt{BC Pilot}, obtained by augmenting real-world teleoperation demonstrations; the bottom row shows training data for \texttt{Expert Pilot}, collected from successful rollouts of the residual copilot policies. Columns correspond to the three tasks: Gear Meshing, Nut Threading, and Peg Insertion. The top row (left to right) contains 442k, 786k, and 427k transitions, while the bottom row contains 220k, 758k, and 195k transitions.}
    \label{state_diffusion_coverage}
\end{figure*}

\subsubsection{RL Hyperparameters}
\label{sec:rl_hparams}
We train policies with PPO using an actor--critic architecture with a shared recurrent backbone (LSTM) followed by an MLP, and a Gaussian policy head. We apply normalization to observations, value targets, and advantages. Key architecture and PPO hyperparameters are summarized in Table~\ref{tab:rl_hparams}.

\subsection{Simulation Setup}
\subsubsection{Admittance Control}
We deploy the same task-space admittance model in simulation as in Sec.~\ref{eq:admittance}. At each control step, we represent the end-effector pose as $\xi\in\mathbb{R}^6$ (position and axis--angle orientation) with reference $\xi_{\mathrm{ref}}$, and compute the pose error $e = \xi - \xi_{\mathrm{ref}}$. We maintain the task-space velocity $\dot{\xi}\in\mathbb{R}^6$ and use diagonal virtual parameters $(M,D,K)$ to form the spring--damper wrench
\[
f_{\mathrm{sd}} = D\dot{\xi} + Ke.
\]
Given the measured external wrench $f$, we integrate the admittance dynamics with forward Euler:
\[
\ddot{\xi} = M^{-1}\!\left(f - f_{\mathrm{sd}}\right),\quad
\dot{\xi} \leftarrow \dot{\xi} + \Delta t\,\ddot{\xi},\quad
\xi \leftarrow \xi + \Delta t\,\dot{\xi}.
\]
We update orientation by converting the angular component of $\Delta t\,\dot{\xi}$ to an incremental axis--angle rotation and applying it via quaternion composition.

The task-space admittance control and joint-space PD parameters can be found in Table ~\ref{tab:controller_params}. 

\subsubsection{Task Progression Metric}
We define task progression as
$1 - {e}/{e_{\max}}$,
where error $e$ is clipped to $[0, e_{\max}]$ so that the metric lies in $[0, 1]$. A value of 1 indicates task completion (zero error), while 0 indicates no meaningful progress. For Gear Meshing and Peg Insertion, error is the Euclidean distance between the end-effector pose and a pre-defined success pose, with $\text{e}_{\max} = 15\,\text{cm}$. For Nut Threading, error is the remaining yaw rotation needed after insertion, with $\text{e}_{\max} = 90°$.

\subsubsection{Physics Engine}
\label{sec:physx}
We use the NVIDIA PhysX engine for rigid-body simulation with a configuration tailored for stable contact-rich manipulation. We employ the Temporal Gauss--Seidel (TGS) solver and set a high number of position iterations (192) to reduce interpenetration and improve contact resolution, while keeping a single velocity iteration for efficiency. To stabilize frictional contact, we use conservative friction thresholds and correlation distances, and set a low bounce threshold velocity (0.2) to suppress spurious rebounds. The GPU solver is provisioned with large contact and collision buffers, and we restrict partitioning to a single partition to ensure deterministic and stable contact dynamics.

\subsection{Behavior Cloning Policy Training}
\subsubsection{State-based Diffusion Policy}
We train state-based Diffusion Policies for \texttt{BC Pilot} and \texttt{Expert Pilot} in \ref{exp1} using a standard DDPM formulation with a fixed action horizon. The policy operates on a low-dimensional state representation \(s_t\) that closely mirrors the residual observation space but excludes action-dependent terms. Specifically, \(s_t\) includes proprioceptive and task-relevant variables (end-effector pose, gripper state, and relative object poses), while omitting the base action and previous residual action, yielding a 20-dimensional state vector.

For \texttt{BC Pilot}, we expand the teleoperation dataset to 2000 episodes by applying global geometric augmentations—random translations and rotations—to entire trajectories. Translations are sampled from \(\mathcal{N}(0,\,0.02)\) and rotations from \(\mathcal{N}(0,\,0.087)\). For \texttt{Expert Pilot}, we collect 2000 successful episodes by rolling out the corresponding \texttt{Residual Copilot} policy. The resulting training data distributions for these state-based diffusion policies are shown in Fig.~\ref{state_diffusion_coverage}.

The diffusion process, noise schedule, and optimization settings are shared across the state- and vision-based variants. The two policies differ only in their input modalities and encoder architectures, while the diffusion decoder and training procedure remain identical. Key diffusion and optimization hyperparameters are summarized in Table~\ref{tab:dp_hparams}.

\begin{figure}[t]
    \centering
    \includegraphics[width=\linewidth]{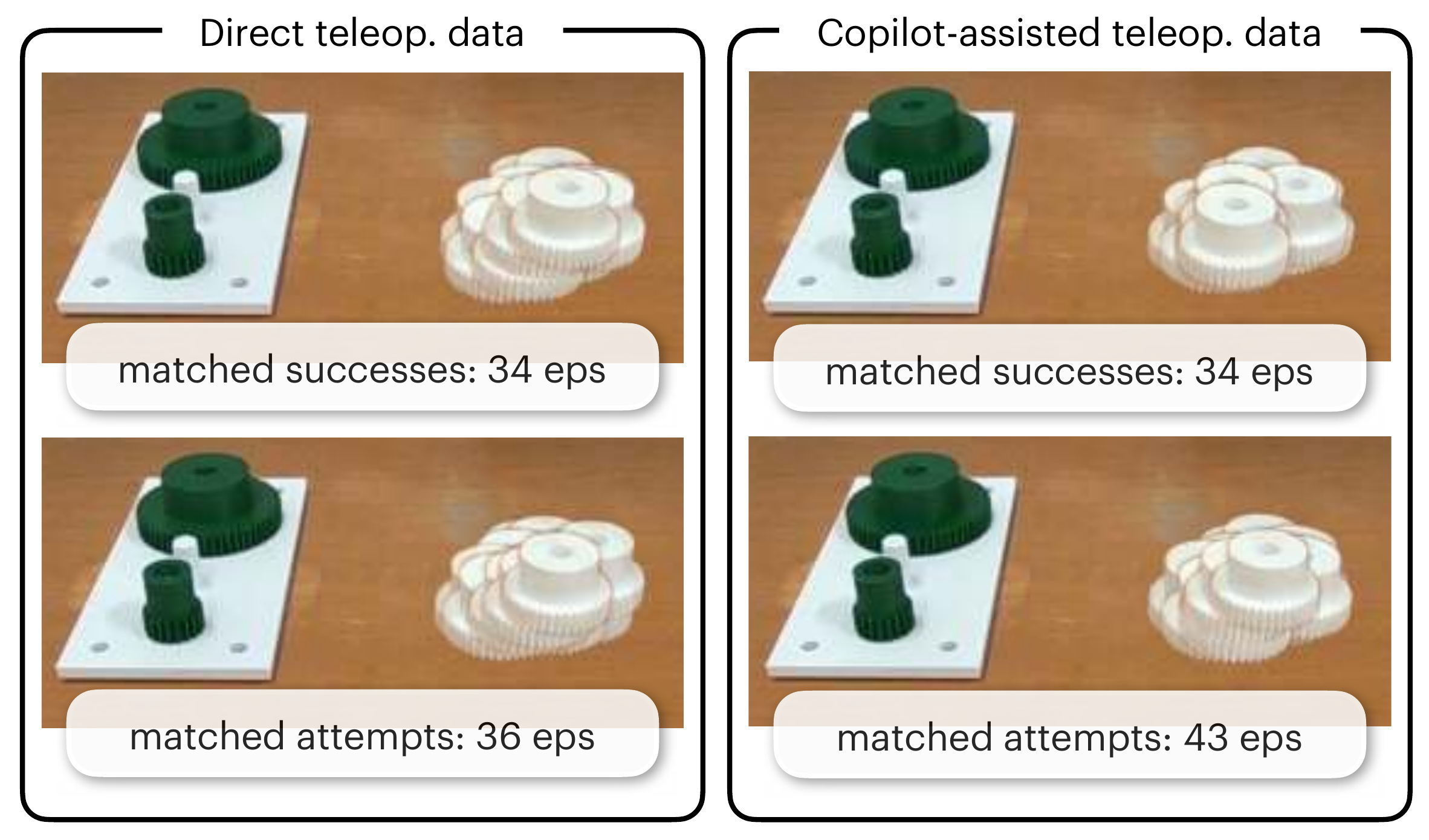}
    \caption{\small \textbf{Vision-based DP training data distributions.}
    Spatial coverage visualization of initial states in the training set used to train vision-based diffusion policies.
    Columns show episodes collected via direct vs.\ copilot-assisted teleoperation. Rows correspond to the two controlled ~\ref{exp3} settings (same number of successful episodes and same number of teleoperation attempts), isolating how data quality and collection efficiency affect downstream policy learning.}
    \label{fig:coverage}
\end{figure}

\subsubsection{Vision-based Diffusion Policy}
We additionally train several vision-based Diffusion Policies for \ref{exp3}, where action-sequence predictions are conditioned on visual observations. In this setting, the state input \(s_t\) is restricted to the robot's proprioceptive state (i.e., end-effector position, orientation, and gripper openness), yielding an 8-dimensional vector. Visual information is provided separately via a single \(200\times200\) RGB image from the front D455 camera. Fig.~\ref{fig:coverage} visualizes the spatial coverage of the initial workspace for each dataset.

RGB observations are encoded using a ResNet-18 backbone with a spatial softmax keypoint layer, whose output is fused with the low-dimensional state before diffusion-based action decoding. Aside from the visual encoder and the reduced state definition, the diffusion model, horizon, and training hyperparameters are shared with the state-based policy. Standard image augmentations are applied during training to improve robustness. All architectural and diffusion hyperparameters common to both variants are reported in Table~\ref{tab:dp_hparams}.

\subsection{User Study Details}
\subsubsection{Participant Information}
We recruited 16 non-author participants (ages 20--30), all male students. Novice teleoperators had limited prior experience, having collected fewer than 20 teleoperation trajectories (often their first exposure to teleoperation). Experienced teleoperators had substantially more prior experience, including prior data collection for imitation learning, and had collected at least 50 teleoperation trajectories.

\subsubsection{Participant Questionnaire}
All subjective metrics were collected using 7-point Likert-scale questions.
Each question followed the same format:
\emph{``How much \textit{X} did you feel when using method \textit{Y}?''}
where \textit{X} denotes the criterion (e.g., mental demand, effort, trust, satisfaction) and \textit{Y} denotes the teleoperation method.
Responses ranged from 1 (least \textit{X}) to 7 (most \textit{X}).
Criteria labeled as \emph{User Demand} are better when lower, while criteria labeled as \emph{User Satisfaction} are better when higher.

\subsubsection{Participant Instructions}
\label{app:instructions}
The full instruction sheet provided to participants prior to the experiment is included on the final page of the appendix (Sec.~\ref{sec:participant_instructions}).

\onecolumn
\section{Participant Instruction Sheet}
\label{sec:participant_instructions}
\textbf{Method Overview.}
In this study, you will perform teleoperation using two different methods:
\begin{enumerate}[leftmargin=3em]
    \item \textbf{Direct Teleoperation.}
    The robot directly mirrors the motion of the teacher arm by matching its end-effector position and orientation. You have full control of the robot and are responsible for completing the entire task from start to finish.

    \item \textbf{Copilot-Assisted Teleoperation.}
    In this mode, a copilot agent assists your teleoperation. Your commands from the teacher arm are combined with corrective actions from the copilot before being executed by the robot. You share control with the copilot: you should focus on high-level, coarse motions, while the copilot corrects local position and orientation errors. For insertion tasks, the copilot will complete the insertion once the robot is near the insertion region. For screwing tasks, the copilot will amplify your rotations and regulate contact force.
\end{enumerate}

\smallskip
\textbf{Teleoperation Objective.}
Your teleoperation objective has three priorities:
\begin{enumerate}[leftmargin=3em]
    \item \textbf{Safety and contact regulation.}
    Always prioritize safe interaction. Noticeable shakiness during contact indicates excessive force and should be avoided.
    \item \textbf{Task success.}
    Try to successfully complete the task to the best of your ability.
    \item \textbf{Input smoothness.}
    Aim for smooth motions. Avoid sudden accelerations and excessively fast movements.
\end{enumerate}

\smallskip
\textbf{Task Descriptions.}
You will perform one of the following tasks:
\begin{itemize}[leftmargin=3em]
    \item \textbf{Gear Meshing:} Pick up the medium gear, insert it onto its shaft, and mesh it with the other gears. A trial is successful only if the gear remains grasped until insertion is fully completed.
    \item \textbf{Nut Threading:} Pick up an M32 nut and tighten it onto the bolt by at least $180^\circ$. Due to hardware limits, this is performed as three consecutive $60^\circ$ turns.
    \item \textbf{Peg Insertion:} Pick up an $8\,\mathrm{mm}$-long peg and insert it into the base. A trial is successful only if the peg reaches the bottom of the receptacle.
\end{itemize}

For all tasks, failure is defined as either dropping the object before success or exceeding the xArm safety force threshold.

\smallskip
\textbf{Experiment Procedure.}
Before the experiment begins, you will have 5 minutes to familiarize yourself with each teleoperation method. During the experiment, you will alternate between direct teleoperation and copilot-assisted teleoperation. At the start of each trial, you will be told which method to use.

At the beginning of every trial, place the teacher arm on the mount to ensure a consistent initial pose. After the experimenter says ``start,'' you may move the arm off the mount and begin teleoperating. The experimenter will say ``stop'' when the trial ends due to either success or failure, and then the next trial will begin.

You will perform only one task in the study. The number of trials per method is:
\begin{itemize}[leftmargin=3em]
    \item \textbf{Gear meshing}: 10 trials per method.
    \item \textbf{Peg insertion}: 8 trials per method.
    \item \textbf{Nut threading}: 5 trials per method.
\end{itemize}

After completing all trials, you will fill out a questionnaire about your experience with each method, including workload and satisfaction.

\end{document}